\documentclass[conference]{IEEEtran}
\usepackage{cite}
\usepackage{amsmath,amssymb,amsfonts}
\usepackage{algorithmic}
\usepackage{graphicx}
\usepackage{textcomp}
\usepackage{array}
\usepackage{mathrsfs}
\usepackage{graphicx}
\usepackage{mathtools}
\def\BibTeX{{\rm B\kern-.05em{\sc i\kern-.025em b}\kern-.08em
    T\kern-.1667em\lower.7ex\hbox{E}\kern-.125emX}}
\begin{document}

\title{Deep Quaternion Networks}

\author{\IEEEauthorblockN{Chase J. Gaudet}
\IEEEauthorblockA{\textit{School of Computing \& Informatics} \\
\textit{University of Lousiana at Lafayette}\\
Lafayette, USA \\
cjg7182@louisiana.edu}
\and
\IEEEauthorblockN{Anthony S. Maida}
\IEEEauthorblockA{\textit{School of Computing \& Informatics} \\
\textit{University of Lousiana at Lafayette}\\
Lafayette, USA \\
maida@louisiana.edu}
}

\maketitle

\begin{abstract}
The field of deep learning has seen significant advancement in recent years.
However, much of the existing work has been focused on real-valued numbers.
Recent work has shown that a deep learning system using the complex numbers can be deeper for a fixed parameter budget compared to its real-valued counterpart.
In this work, we explore the benefits of generalizing one step further into the hyper-complex numbers, quaternions specifically, and provide the architecture components needed to build deep quaternion networks.
We develop the theoretical basis by reviewing quaternion convolutions, developing a novel quaternion weight initialization scheme, and developing novel algorithms for quaternion batch-normalization.
These pieces are tested in a classification model by end-to-end training on the CIFAR-10 and CIFAR-100 data sets and a segmentation model by end-to-end training on the KITTI Road Segmentation data set. 
These quaternion networks show improved convergence compared to real-valued and complex-valued networks, especially on the segmentation task, while having fewer parameters.
\end{abstract}


\section{Introduction}
There have been many advances in deep neural network architectures in the past few years.
One such improvement is a normalization technique called batch normalization \cite{ioffe2015batch} that standardizes the activations of layers inside a network using minibatch statistics.
It has been shown to regularize the network as well as provide faster and more stable training.
Another improvement comes from architectures that add so called shortcut paths to the network.
These shortcut paths connect later layers to earlier layers typically, which allows for the stronger gradients to propagate to the earlier layers.
This method can be seen in Highway Networks \cite{srivastava2015training} and Residual Networks  \cite{he2016deep}.
Other work has been done to find new activation functions with more desirable properties.
One example is the exponential linear unit (ELU) \cite{clevert2015fast}, which attempts to keep activations standardized.
All of the above methods are combating the vanishing gradient problem \cite{hochreiter1991untersuchungen} that plagues deep architectures.
With solutions to this problem appearing it is only natural to move to a system that will allow one to construct deeper architectures with as low a parameter cost as possible.

Other work in this area has explored the use of complex and hyper-complex numbers, which are a generalization of the complex, such as quaternions.
Using complex numbers in recurrent neural networks (RNNs) has been shown to increase learning speed and provide a more noise robust memory retrieval mechanism \cite{arjovsky2016unitary, danihelka2016associative, wisdom2016full}.
The first formulation of complex batch normalization and complex weight initialization is presented by \cite{trabelsi2017deep} where they achieve some state of the art results on the MusicNet data set.
Hyper-complex numbers are less explored in neural networks, but have seen use in manual image and signal processing techniques \cite{bulow1999hypercomplex, sangwine2000colour, bulow2001hypercomplex}.
Examples of using quaternion values in networks is mostly limited to architectures that take in quaternion inputs or predict quaternion outputs, but do not have quaternion weight values \cite{rishiyur2006neural, kendall2015posenet}. 
There are some more recent examples of building models that use quaternions represented as real-values. 
In \cite{parcollet2016quaternion} they used a quaternion multi-layer perceptron (QMLP) for document understanding and \cite{minemoto2017feed} uses a similar approach in processing multi-dimensional signals. 

Building on \cite{trabelsi2017deep} our contribution in this paper is to formulate and implement quaternion convolution, batch normalization, and weight initialization \footnote{Source code located at https://github.com/gaudetcj/DeepQuaternionNetworks}.
There arises some difficulty over complex batch normalization that we had to overcome as their is no analytic form for our inverse square root matrix.

\section{Motivation and Related Work}
The ability of quaternions to effectively represent spatial transformations and analyze multi-dimensional signals makes them promising for applications in artificial intelligence.

One common use of quaternions is for representing rotation into a more compact form. 
PoseNet \cite{kendall2015posenet} used a quaternion as the target output in their model where the goal was to recover the $6-$DOF camera pose from a single RGB image.
The ability to encode rotations may make a quaternion network more robust to rotational variance.

Quaternion representation has also been used in signal processing.  
The amount of information in the phase of an image has been shown to be sufficient to recover the majority of information encoded in its magnitude by Oppenheim and Lin \cite{oppenheim1981importance}.
The phase also encodes information such as shapes, edges, and orientations.
Quaternions can be represented as a 2~x~2 matrix of complex numbers, which gives them a group of phases potentially holding more information compared to a single phase.

Bulow and Sommer \cite{bulow2001hypercomplex} used the higher complexity representation of quaternions by extending Gabor's complex signal to a quaternion one which was then used for texture segmentation.
Another use of quaternion filters is shown in \cite{sangwine2000colour} where they introduce a new class of filter based on convolution with hyper-complex masks, and present three color edge detecting filters. 
These filters rely on a three-space rotation about the grey line of RGB space and when applied to a color image produce an almost greyscale image with color edges where the original image had a sharp change of color.
More quaternion filter use is shown in \cite{shi2007quaternion} where they show that it is effective in the context of segmenting color images into regions of similar color texture. 
They state the advantage of using quaternion arithmetic is that a color can be represented and analyzed as a single entity (by assigning each color channel to an imaginary axis), which we will see holds for quaternion convolution in a convolutional neural network architecture as well in Section \ref{s:qc}.

A quaternionic extension of a feed forward neural network, for processing multi-dimensional signals, is shown in \cite{minemoto2017feed}.
They expect that quaternion neurons operate on multi-dimensional signals as single entities, rather than real-valued neurons that deal with each element of signals independently.
A convolutional neural network (CNN) should be able to learn a powerful set of quaternion filters for more impressive tasks.

Another large motivation is discussed in \cite{trabelsi2017deep}, which is that complex numbers are more efficient and provide more robust memory mechanisms compared to the reals \cite{bulow1999hypercomplex, sangwine2000colour, bulow2001hypercomplex}.
They continue that residual networks have a similar architecture to associative memories since the residual shortcut paths compute their residual and then sum it into the memory provided by the identity connection.
Again, given that quaternions can be represented as a complex group, they may provide an even more efficient and robust memory mechanisms.

\section{Quaternion Network Components}
This section will include the work done to obtain a working deep quaternion network. 
Some of the longer derivations are given in the Appendix.

\subsection{Quaternion Representation}
In 1833 Hamilton proposed complex numbers $\mathbb{C}$ be defined as the set $\mathbb{R}^2$ of ordered pairs $(a, b)$ of real numbers.
He then began working to see if triplets $(a,b,c)$ could extend multiplication of complex numbers.
In 1843 he discovered a way to multiply in four dimensions instead of three, but the multiplication lost commutativity.
This construction is now known as quaternions.
Quaternions are composed of four components, one real part, and three imaginary parts.
Typically denoted as
\begin{equation}
\mathbb{H} = \{a + b\textit{i} + c\textit{j} + d\textit{k}~:~a,b,c,d \in \mathbb{R}\}
\label{eq:quaternion1}
\end{equation}
where $a$ is the real part, $(i,j,k)$ denotes the three imaginary axis, and $(b,c,d)$ denotes the three imaginary components.
Quaternions are governed by the following arithmetic:
\begin{equation}
i^2=j^2=k^2=ijk=-1
\label{eq:quarternion2}
\end{equation}
which, by enforcing distributivity, leads to the noncommutative multiplication rules
\begin{equation}
ij=k,~jk=i,~ki=j,~ji=-k,~kj=-i,~ik=-j
\label{eq:quarternion3}
\end{equation}

Since we will be performing quaternion arithmetic using reals it is useful to embed $\mathbb{H}$ into a real-valued representation.
There exists an injective homomorphism from $\mathbb{H}$ to the matrix ring $M(4,\mathbb{R})$ where $M(4,\mathbb{R})$ is a 4x4 real matrix.
The 4~x~4 matrix can be written as
\begin{align}
\begin{bmatrix}
 a & -b & -c & -d \\ 
 b & a & -d & c \\
 c & d & a & -b \\
 d & -c & b & a 
\end{bmatrix}= &~~a
\begin{bmatrix}
 1 & 0 & 0 & 0 \\ 
 0 & 1 & 0 & 0 \\
 0 & 0 & 1 & 0 \\
 0 & 0 & 0 & 1 
\end{bmatrix}
\nonumber \\ &+ b 
\begin{bmatrix}
 0 & -1 & 0 & 0 \\ 
 1 & 0 & 0 & 0 \\
 0 & 0 & 0 & -1 \\
 0 & 0 & 1 & 0 
\end{bmatrix}
\nonumber \\ &+ c
\begin{bmatrix}
 0 & 0 & -1 & 0 \\ 
 0 & 0 & 0 & 1 \\
 1 & 0 & 0 & 0 \\
 0 & -1 & 0 & 0 
\end{bmatrix}
\nonumber \\ &+ d
\begin{bmatrix}
 0 & 0 & 0 & -1 \\ 
 0 & 0 & -1 & 0 \\
 0 & 1 & 0 & 0 \\
 1 & 0 & 0 & 0 
\end{bmatrix}.
\label{eq:m4r}
\end{align}
This representation of quaternions is not unique, but we will stick to the above in this paper.
It is also possible to represent $\mathbb{H}$ as $M(2,\mathbb{C})$ where $M(2,\mathbb{C})$ is a 2~x~2 complex matrix.

With our real-valued representation a quaternion real-valued $2D$ convolution layer can be expressed as follows. 
Say that the layer has $N$ feature maps such that $N$ is divisible by 4.
We let the first $N/4$ feature maps represent the real components, the second $N/4$ represent the $i$ imaginary components, the third $N/4$ represent the $j$ imaginary components, and the last $N/4$ represent the $k$ imaginary components.

\subsection{Quaternion Differentiability}
In order for the network to perform backpropagation the cost function and activation functions used must be differentiable with respect to the real, $i$, $j$, and $k$ components of each quaternion parameter of the network.
As the complex chain rule is shown in \cite{trabelsi2017deep}, we provide the quaternion chain rule which is given in the Appendix section \ref{a:diff}.

\subsection{Quaternion Convolution}\label{s:qc}
Convolution in the quaternion domain is done by convolving a quaternion filter matrix $\textbf{W}=\textbf{A}+\textit{i}~\textbf{B}+\textit{j}~\textbf{C}+\textit{k}~\textbf{D}$ by a quaternion vector $\textbf{h}=\textbf{w}+\textit{i}~\textbf{x}+\textit{j}~\textbf{y}+\textit{k}~\textbf{z}$.
Here $\textbf{A}$, $\textbf{B}$, $\textbf{C}$, and $\textbf{D}$ are real-valued matrices and $\textbf{w}$, $\textbf{x}$, $\textbf{y}$, and $\textbf{z}$ are real-valued vectors.
Performing the convolution by using the distributive property and grouping terms one gets
\begin{align}
\textbf{W}\ast \textbf{h} = &~(\textbf{A}\ast\textbf{w}-\textbf{B}\ast\textbf{x}-\textbf{C}\ast\textbf{y}-\textbf{D}\ast\textbf{z}) + \nonumber \\ 
&\textit{i}(\textbf{A}\ast\textbf{x}+\textbf{B}\ast\textbf{w}+\textbf{C}\ast\textbf{z}-\textbf{D}\ast\textbf{y}) + \nonumber \\
&\textit{j}(\textbf{A}\ast\textbf{y}-\textbf{B}\ast\textbf{z}+\textbf{C}\ast\textbf{w}+\textbf{D}\ast\textbf{x}) + \nonumber \\
&\textit{k}(\textbf{A}\ast\textbf{z}+\textbf{B}\ast\textbf{y}-\textbf{C}\ast\textbf{x}+\textbf{D}\ast\textbf{w}).
\end{align}
Using a matrix to represent the components of the convolution we have:
\begin{equation}
\begin{bmatrix}
 \mathscr{R}(\textbf{W}\ast \textbf{h}) \\ 
 \mathscr{I}(\textbf{W}\ast \textbf{h}) \\
 \mathscr{J}(\textbf{W}\ast \textbf{h}) \\
 \mathscr{K}(\textbf{W}\ast \textbf{h}) 
\end{bmatrix}
=
\begin{bmatrix}
 \textbf{A} & -\textbf{B} & -\textbf{C} & -\textbf{D}\\
 \textbf{B} & \textbf{A} & -\textbf{D} & \textbf{C} \\
 \textbf{C} & \textbf{D} & \textbf{A} & -\textbf{B} \\
 \textbf{D} & -\textbf{C} & \textbf{B} & \textbf{A} \\
\end{bmatrix}
\ast
\begin{bmatrix}
 \textbf{w} \\ 
 \textbf{x} \\
 \textbf{y} \\
 \textbf{z}
\end{bmatrix}
\label{eq:qconvolve2}
\end{equation}

\begin{figure*}
	\centering
		\includegraphics[width=1.0\textwidth]{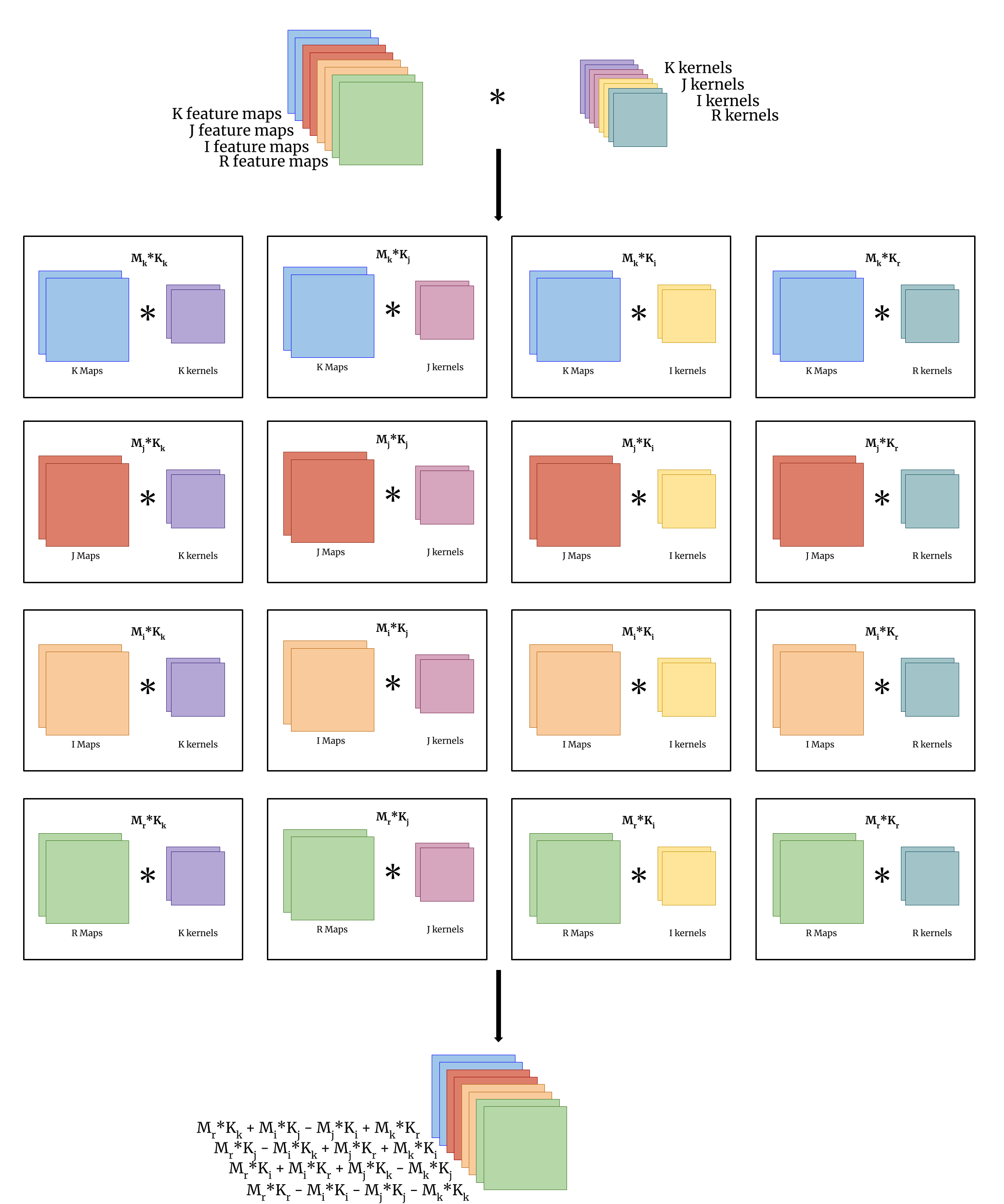}
	\caption{An illustration of quaternion convolution.}
	\label{f:quatconv}
\end{figure*}

An example is shown in Fig.~\ref{f:quatconv}, which is useful to visualize one of the main motivational factors of quaternions for CNNs.
Notice that the result of the quaternion convolution produces a unique linear combination of each axis per the result of a single axis.
This comes from the structure of quaternion multiplication and is forcing each axis of the kernel to interact with each axis of the image.
Real-valued convolution simply multiplies each channel of the kernel with the corresponding channel of the image.
The quaternion convolution is similar to a mixture of standard convolution and depthwise separable convolution from \cite{chollet2016xception}. 
Depthwise separable convolution is where first a flat convolution kernel (no depth to match the depth of the feature image) is applied separately to each feature map.
This is only giving spatial context on each feature map individually.
Then a $1\times1$ convolution is applied to the results of the previous operation to get a linear interaction of the feature maps, projecting them into a new feature map space.

The quaternion network's reuse of filters on every axis and combination may help extract texture information across channels as seen in \cite{shi2007quaternion}.
One can think in terms of a RGB image where the greyscale of the image can be the real axis and the RGB channels individually can be the $i, j, k$ axes.
Then a quaternion kernel convolved against this quaternion image will view the colors as a single entity, unlike standard real-valued convolution.
Since a quaternion can be thought of as a vector, the quaternion kernels and feature maps can be thought of as vectors as well.

\subsection{Quaternion Batch-Normalization}
Batch-normalization \cite{ioffe2015batch} is used by the vast majority of all deep networks to stabilize and speed up training.
It works by keeping the activations of the network at zero mean and unit variance.
The original formulation of batch-normalization only works for real-values. 
Applying batch normalization to complex or hyper-complex numbers is more difficult, one can not simply translate and scale them such that their mean is 0 and their variance is 1.
This would not give equal variance in the multiple components of a complex or hyper-complex number.
To overcome this for complex numbers a whitening approach is used \cite{trabelsi2017deep}, which scales the data by the square root of their variances along each of the two principle components.
We use the same approach, but must whiten 4D vectors.

However, an issue arises in that there is no nice way to calculate the inverse square root of a 4~x~4 matrix.
It turns out that the square root is not necessary and we can instead use the Cholesky decomposition on our covariance matrix.
The details of why this works for whitening given in the Appendix section \ref{a:whitening}.
Now our whitening is accomplished by multiplying the \textbf{0}-centered data ($\textbf{x} - \mathbb{E}[\textbf{x}]$) by \textbf{W}:
\begin{equation}
\tilde{x} = \textbf{W}(\textbf{x} - \mathbb{E}[\textbf{x}])
\label{eq:white4d}
\end{equation}
where \textbf{W} is one of the matrices from the Cholesky decomposition of $\textbf{V}^{-1}$ where \textbf{V} is the covariance matrix given by:
\begin{align*}
\textbf{V}
=&
\begin{bmatrix}
 V_{rr} & V_{ri} & V_{rj} & V_{rk} \\
 V_{ir} & V_{ii} & V_{ij} & V_{ik} \\
 V_{jr} & V_{ji} & V_{jj} & V_{jk} \\
 V_{kr} & V_{ki} & V_{kj} & V_{kk}
\end{bmatrix}
\label{eq:V4d}
\end{align*}
where each $V$ is the covariance between its two subscripts which represent the real, $i$, $j$, and $k$ components of $\textbf{x}$ respectively.

Real-valued batch normalization also uses two learned parameters, $\beta$ and $\gamma$. 
Our shift parameter {\boldmath$\beta$} must shift a quaternion value so it is a quaternion value itself with real, $i$, $j$, and $k$ as learnable components. 
The scaling parameter {\boldmath$\gamma$} is a symmetric matrix of size matching $\textbf{V}$ given by:
\begin{equation}
\mathbf{\gamma}
=
\left( 
\begin{array}{cccc}
\gamma_{rr} & \gamma_{ri} & \gamma_{rj} & \gamma_{rk} \\
\gamma_{ri} & \gamma_{ii} & \gamma_{ij} & \gamma_{ik} \\
\gamma_{rj} & \gamma_{ij} & \gamma_{jj} & \gamma_{jk} \\
\gamma_{rk} & \gamma_{ik} & \gamma_{jk} & \gamma_{kk}
\end{array}
\right)
\label{eq:gamma}
\end{equation}
Because of its symmetry it has only ten learnable parameters. 
The variance of the components of input $\tilde{\textbf{x}}$ are variance 1 so the diagonal of {\boldmath$\gamma$} is initialized to $1/\sqrt{4}$ in order to obtain a modulus of 1 for the variance of the normalized value. 
The off diagonal terms of {\boldmath$\gamma$} and all components of {\boldmath$\beta$} are initialized to 0.
The quaternion batch normalization is defined as:
\begin{equation}
\mbox{BN}(\tilde{\textbf{x}}) = \mathbf{\gamma}\tilde{\textbf{x}} + \mathbf{\beta}
\label{eq:qbn}
\end{equation}

\subsection{Quaternion Weight Initialization}
The proper initialization of weights is vital to convergence of deep networks. 
In this work we derive our quaternion weight initialization using the same procedure as Glorot and Bengio \cite{glorot2010understanding} and He et al. \cite{he2015delving}.

To begin we find the variance of a quaternion weight:
\begin{align}
W = &~|W|e^{(\mbox{cos}\phi_1 \textit{i} + \mbox{cos}\phi_2 \textit{j} + \mbox{cos}\phi_3 \textit{k})\theta} \nonumber \\
= &~\mathscr{R}\{W\} + \mathscr{I}\{W\} + \mathscr{J}\{W\} + \mathscr{K}\{W\}.
\label{eq:quaternion_weight}
\end{align}
where $|W|$ is the magnitude, $\theta$ and $\phi$ are angle arguments, and $\mbox{cos}^2\phi_1 + \mbox{cos}^2\phi_2 + \mbox{cos}^2\phi_3 = 1$ \cite{turner2002}.

Variance is defined as
\begin{equation}
\mbox{Var}(W) = \mathbb{E}[|W|^2] - (\mathbb{E}[W])^2,
\label{eq:variance}
\end{equation}
but since $W$ is symmetric around 0 the term $(\mathbb{E}[W])^2$ is 0. 
We do not have a way to calculate $\mbox{Var}(W) = \mathbb{E}[|W|^2]$ so we make use of the magnitude of quaternion normal values $|W|$, which follows an independent normal distribution with four degrees of freedom (DOFs).
We can then calculate the expected value of $|W|^2$ to find our variance
\begin{equation}
\mathbb{E}[|W|^2] = \int_{-\infty}^\infty x^2 f(x) ~dx = 4\sigma^2
\label{eq:expected}
\end{equation}
where $f(x)$ is the four DOF distribution given in the Appendix.

And since $\mbox{Var}(W) = \mathbb{E}[|W|^2]$, we now have the variance of $W$ expressed in terms of a single parameter $\sigma$:
\begin{equation}
\mbox{Var}(W) = 4\sigma^2.
\label{eq:variance_sigma}
\end{equation}

To follow the Glorot and Bengio \cite{glorot2010understanding} initialization we have $\mbox{Var}(W) = 2/(n_{in}+n_{out})$, where $n_{in}$ and $n_{out}$ are the number of input and output units respectivly. 
Setting this equal to \eqref{eq:variance_sigma} and solving for $\sigma$ gives $\sigma = 1/\sqrt{2(n_{in}+n_{out})}$.
To follow He et al. \cite{he2015delving} initialization that is specialized for rectified linear units (ReLUs) \cite{nair2010rectified}, then we have $\mbox{Var}(W) = 2/n_{in}$, which again setting equal to \eqref{eq:variance_sigma} and solving for $\sigma$ gives $\sigma = 1/\sqrt{2n_{in}}$.

As shown in \eqref{eq:quaternion_weight} the weight has components $|W|$, $\theta$, and $\phi$. 
We can initialize the magnitude $|W|$ using our four DOF distribution defined with the appropriate $\sigma$ based on which initialization scheme we are following. 
The angle components are initialized using the uniform distribution between $-\pi$ and $\pi$ where we ensure the constraint on $\phi$.

\section{Experimental Results}
Our experiments covered image classification using both the CIFAR-10 and CIFAR-100 benchmarks \cite{krizhevsky2009learning} and image segmentation using the KITTI Road Estimation benchmark \cite{Fritsch2013ITSC}. 
The CIFAR datasets are $32\times32$ color images of 10 and 100 classes receptively.
Each image only contains one class and labels are provided.
The KITTI dataset is large color images of varying sizes depicting roads as seen from a driver's perspective.
Each image has a corresponding label image in which each pixel is an integer value relating to a class of road or not road.
We chose CIFAR because it is a extremely common benchmark task making it a good sanity check.
The KITTI dataset was chosen because it is a fairly common, color segmentation benchmark and has binary classes which made it a simple test.
All training was done on a single Nvidia 980Ti.

\subsection{Classification}
We use the same architecture as the large model in \cite{trabelsi2017deep}, which is a 110 layer Residual model similar to the one in \cite{he2016deep}.
There is one difference between the real-valued network and the ones used for both the complex and hyper-complex valued networks.
Because the datasets are all real-valued the network must learn the imaginary or quaternion components.
We use the same technique as \cite{trabelsi2017deep} where there is an additional residual block immediately after the input which will learn the hyper-complex components
\begin{equation*}
BN \rightarrow ReLU \rightarrow Conv \rightarrow BN \rightarrow ReLU \rightarrow Conv.
\end{equation*}
One of these blocks exist per imaginary component and are concatenated with the original input image.
Another possible choice if using color images is to use the gray scale image as the real axis and then use the red, green, and blue channels as the $i, j,$ and $k$ axis respectively.
With this choice it is not necessary to use the above block after input to learn the imaginary components.

To maintain the same parameter budget among the three network types we divided the number of filters per layer of the real network by a factor of two for the complex, and by a factor of four for the quaternion.

The architecture for all models consists of 3 stages of repeating residual blocks,
\begin{equation*}
BN \rightarrow ReLU \rightarrow Conv \rightarrow BN \rightarrow ReLU \rightarrow Conv,
\end{equation*}
where at the end of each stage the images are downsized by a strided convolution.
For these classification experiments we ran a shallow network where the stages contained 2, 1, and 1 residual blocks respectively and a deep network where the stages contained 10, 9, and 9 residual blocks respectively.
Each stage also doubles the previous stage's number of convolution kernels.
For example the real model has 32 kernels in the first stage, 64 in the second, and finally 124 in the last.
The last two layers are a global average pooling layer followed by a single fully connected layer with a softmax function used to classify the input as either one of the 10 classes in CIFAR-10 or one of the 100 classes in CIFAR-100.

We also followed their training procedure of using the backpropagation algorithm with Stochastic Gradient Descent with Nesterov momentum \cite{nesterov1983method} set at 0.9.
The norm of the gradients were clipped to 1 and a custom learning rate scheduler was used.
The learning scheduler was the same used in \cite{trabelsi2017deep} for a direct comparison in performance.
The learning rate was initially set to 0.01 for the first 10 epochs and then set to 0.1 from epoch 11-100 and then cut by a factor of 10 at epochs 120 and 150.
Table~\ref{t:results1} presents our results alongside the real and complex valued networks.
Our quaternion models outperform the real and complex networks on both datasets with a smaller parameter count.
The quaternion models do take roughly 50\% longer to train due to the computationally intense operations of quaternion batch normalization.

\begin{table}[h]
	\centering
		\begin{tabular}{l c c c c}
			\hline
			Architecture & Params & CIFAR-10 & CIFAR-100 \\
			\hline
			Shallow Real & 508,932 & 6.82 & 32.02 \\
			Shallow Complex & 257,412 & 6.91 & 31.65 \\
			Shallow Quaternion & 133,560 & \textbf{6.77} & \textbf{30.59} \\
			\hline
			Deep Real & 3,619,844 & 6.37 & 28.07 \\
			Deep Complex & 1,823,620 & 5.60 & 27.09 \\
			Deep Quaternion & 932,792 & \textbf{5.44} & \textbf{26.01} \\
			\hline
		\end{tabular}
	\caption{Classification error on CIFAR-10 and CIFAR-100. Params is the total number of parameters.}
	\label{t:results1}
\end{table}

\subsection{Segmentation}
For this experiment we used the same model as the above, but cut the number of residual blocks out of the model for memory reasons given that the KITTI data is large color images about $1200 \times 375$ pixels in size.
We only use one model due to resource limitations for this experiment.
It is the same as the small model from the classification experiments which has 2, 1, and 1 blocks at the three stages, but it does not perform any strided convolutions.
It also does not have the global average pooling layer or the fully connected layer.

The last layer is a $1 \times 1$ convolution with a sigmoid output so we are getting a heatmap prediction the same size as the input.
The training procedure is also as above, but the learning rate is scheduled differently.
Here we begin at 0.01 for the first 10 epochs and then set it to 0.1 from epoch 11-50 and then cut by a factor of 10 at 100 and 150.
Table~\ref{t:results2} presents our results along side the real and complex valued networks where we used Intersection over Union (IOU) for performance measure.
Quaternion outperformed the other two by a larger margin compared to the classification tasks and again, with a smaller parameter count.

\begin{table}[h]
	\centering
		\begin{tabular}{l c c}
			\hline
			Architecture & Params & KITTI \\
			\hline
			Real & 507,029 & 0.747 \\
			Complex & 254,037 & 0.769 \\
			Quaternion & 128,701 & \textbf{0.827}
		\end{tabular}
	\caption{IOU on KITTI Road Estimation benchmark.}
	\label{t:results2}
\end{table}

\section{Conclusions}
We have extended upon work looking into complex valued networks by exploring quaternion values.
We presented the building blocks required to build and train deep quaternion networks and used them to test residual architectures on two common image classification benchmarks.
We show that they have competitive performance by beating both the real and complex valued networks with less parameters.
Future work will be needed to test quaternion networks for more segmentation datasets and for audio processing tasks.

\section{Acknowledgment}
We would like to thank James Dent of the University of Louisiana at Lafayette Physics Department for helpful discussions.
We also thank Fugro for research time on this project.

\bibliography{bib}{}

\begin{thebibliography}{10}

\bibitem{ioffe2015batch}
S.~Ioffe and C.~Szegedy, ``Batch normalization: Accelerating deep network
  training by reducing internal covariate shift,'' {\em arXiv preprint
  arXiv:1502.03167}, 2015.

\bibitem{srivastava2015training}
R.~K. Srivastava, K.~Greff, and J.~Schmidhuber, ``Training very deep
  networks,'' in {\em Advances in neural information processing systems},
  pp.~2377--2385, 2015.

\bibitem{he2016deep}
K.~He, X.~Zhang, S.~Ren, and J.~Sun, ``Deep residual learning for image
  recognition,'' in {\em Proceedings of the IEEE conference on computer vision
  and pattern recognition}, pp.~770--778, 2016.

\bibitem{clevert2015fast}
D.-A. Clevert, T.~Unterthiner, and S.~Hochreiter, ``Fast and accurate deep
  network learning by exponential linear units (elus),'' {\em arXiv preprint
  arXiv:1511.07289}, 2015.

\bibitem{hochreiter1991untersuchungen}
S.~Hochreiter, ``Untersuchungen zu dynamischen neuronalen netzen,'' {\em
  Diploma, Technische Universit{\"a}t M{\"u}nchen}, vol.~91, 1991.

\bibitem{arjovsky2016unitary}
M.~Arjovsky, A.~Shah, and Y.~Bengio, ``Unitary evolution recurrent neural
  networks,'' in {\em International Conference on Machine Learning},
  pp.~1120--1128, 2016.

\bibitem{danihelka2016associative}
I.~Danihelka, G.~Wayne, B.~Uria, N.~Kalchbrenner, and A.~Graves, ``Associative
  long short-term memory,'' {\em arXiv preprint arXiv:1602.03032}, 2016.

\bibitem{wisdom2016full}
S.~Wisdom, T.~Powers, J.~Hershey, J.~Le~Roux, and L.~Atlas, ``Full-capacity
  unitary recurrent neural networks,'' in {\em Advances in Neural Information
  Processing Systems}, pp.~4880--4888, 2016.

\bibitem{trabelsi2017deep}
C.~Trabelsi, O.~Bilaniuk, D.~Serdyuk, S.~Subramanian, J.~F. Santos, S.~Mehri,
  N.~Rostamzadeh, Y.~Bengio, and C.~J. Pal, ``Deep complex networks,'' {\em
  arXiv preprint arXiv:1705.09792}, 2017.

\bibitem{bulow1999hypercomplex}
T.~B{\"u}low, {\em Hypercomplex spectral signal representations for the
  processing and analysis of images}.
\newblock Universit{\"a}t Kiel. Institut f{\"u}r Informatik und Praktische
  Mathematik, 1999.

\bibitem{sangwine2000colour}
S.~J. Sangwine and T.~A. Ell, ``Colour image filters based on hypercomplex
  convolution,'' {\em IEE Proceedings-Vision, Image and Signal Processing},
  vol.~147, no.~2, pp.~89--93, 2000.

\bibitem{bulow2001hypercomplex}
T.~Bulow and G.~Sommer, ``Hypercomplex signals-a novel extension of the
  analytic signal to the multidimensional case,'' {\em IEEE Transactions on
  signal processing}, vol.~49, no.~11, pp.~2844--2852, 2001.

\bibitem{rishiyur2006neural}
A.~Rishiyur, ``Neural networks with complex and quaternion inputs,'' {\em arXiv
  preprint cs/0607090}, 2006.

\bibitem{kendall2015posenet}
A.~Kendall, M.~Grimes, and R.~Cipolla, ``Posenet: A convolutional network for
  real-time 6-dof camera relocalization,'' in {\em Proceedings of the IEEE
  international conference on computer vision}, pp.~2938--2946, 2015.

\bibitem{parcollet2016quaternion}
T.~Parcollet, M.~Morchid, P.-M. Bousquet, R.~Dufour, G.~Linar{\`e}s, and
  R.~De~Mori, ``Quaternion neural networks for spoken language understanding,''
  in {\em Spoken Language Technology Workshop (SLT), 2016 IEEE}, pp.~362--368,
  IEEE, 2016.

\bibitem{minemoto2017feed}
T.~Minemoto, T.~Isokawa, H.~Nishimura, and N.~Matsui, ``Feed forward neural
  network with random quaternionic neurons,'' {\em Signal Processing},
  vol.~136, pp.~59--68, 2017.

\bibitem{oppenheim1981importance}
A.~V. Oppenheim and J.~S. Lim, ``The importance of phase in signals,'' {\em
  Proceedings of the IEEE}, vol.~69, no.~5, pp.~529--541, 1981.

\bibitem{shi2007quaternion}
L.~Shi and B.~Funt, ``Quaternion color texture segmentation,'' {\em Computer
  Vision and image understanding}, vol.~107, no.~1, pp.~88--96, 2007.

\bibitem{chollet2016xception}
F.~Chollet, ``Xception: Deep learning with depthwise separable convolutions,''
  {\em arXiv preprint arXiv:1610.02357}, 2016.

\bibitem{glorot2010understanding}
X.~Glorot and Y.~Bengio, ``Understanding the difficulty of training deep
  feedforward neural networks,'' in {\em Proceedings of the Thirteenth
  International Conference on Artificial Intelligence and Statistics},
  pp.~249--256, 2010.

\bibitem{he2015delving}
K.~He, X.~Zhang, S.~Ren, and J.~Sun, ``Delving deep into rectifiers: Surpassing
  human-level performance on imagenet classification,'' in {\em Proceedings of
  the IEEE International Conference on Computer Vision}, pp.~1026--1034, 2015.

\bibitem{turner2002}
R.~Piziak and D.~Turner, ``The polar form of a quaternion,'' 2002.

\bibitem{nair2010rectified}
V.~Nair and G.~E. Hinton, ``Rectified linear units improve restricted boltzmann
  machines,'' in {\em Proceedings of the 27th International Conference on
  Machine Learning (ICML-10)}, pp.~807--814, 2010.

\bibitem{krizhevsky2009learning}
A.~Krizhevsky and G.~Hinton, ``Learning multiple layers of features from tiny
  images,'' 2009.

\bibitem{Fritsch2013ITSC}
J.~Fritsch, T.~Kuehnl, and A.~Geiger, ``A new performance measure and
  evaluation benchmark for road detection algorithms,'' in {\em International
  Conference on Intelligent Transportation Systems (ITSC)}, 2013.

\bibitem{nesterov1983method}
Y.~Nesterov, ``A method of solving a convex programming problem with
  convergence rate o (1/k2),''

\bibitem{kessy2017optimal}
A.~Kessy, A.~Lewin, and K.~Strimmer, ``Optimal whitening and decorrelation,''
  {\em The American Statistician}, no.~just-accepted, 2017.

\end{thebibliography}
\bibliographystyle{ieeetr}

\section{Appendix}
\subsection{The Generalized Quaternion Chain Rule for a Real-Valued Function}\label{a:diff}
We start by specifying the Jacobian.
Let $L$ be a real valued loss function and $q$ be a quaternion variable such that $q = a+\textit{i}~b+\textit{j}~c+\textit{k}~d$ where $a,b,c,d \in \mathbb{R}$ then,
\begin{align}
\nabla_L(q) &= \frac{\partial L}{\partial q} = \frac{\partial L}{\partial a} + \textit{i}~\frac{\partial L}{\partial b} + \textit{j}~\frac{\partial L}{\partial c} + \textit{k}~\frac{\partial L}{\partial d} \\ \nonumber
&= \frac{\partial L}{\partial \mathbb{R}(q)} + \textit{i}~\frac{\partial L}{\partial \mathbb{I}(q)} + \textit{j}~\frac{\partial L}{\partial \mathbb{J}(q)} + \textit{k}~\frac{\partial L}{\partial \mathbb{K}(q)} \\ \nonumber
&= \mathbb{R}(\nabla_L(q)) + i~\mathbb{I}(\nabla_L(q)) + j~\mathbb{J}(\nabla_L(q)) + k~\mathbb{K}(\nabla_L(q)) 
\label{eq:diff1}
\end{align}
Now let $g = m+\textit{i}~n+\textit{j}~o+\textit{k}~p$ be another quaternion variable where $q$ can be expressed in terms of $g$ and $m,n,o,p \in \mathbb{R}$ we then have,
\begin{align}
\nabla_L(q) &= \frac{\partial L}{\partial g} = \frac{\partial L}{\partial m} + \textit{i}~\frac{\partial L}{\partial n} + \textit{j}~\frac{\partial L}{\partial o} + \textit{k}~\frac{\partial L}{\partial p} \\ \nonumber
&= \frac{\partial L}{\partial a}\frac{\partial a}{\partial m} + \frac{\partial L}{\partial b}\frac{\partial b}{\partial m} + \frac{\partial L}{\partial c}\frac{\partial c}{\partial m} + \frac{\partial L}{\partial d}\frac{\partial d}{\partial m} \\ \nonumber
&~~+ \textit{i}~\left( \frac{\partial L}{\partial a}\frac{\partial a}{\partial n} + \frac{\partial L}{\partial b}\frac{\partial b}{\partial n} + \frac{\partial L}{\partial c}\frac{\partial c}{\partial n} + \frac{\partial L}{\partial d}\frac{\partial d}{\partial n} \right) \\ \nonumber
&~~+ \textit{j}~\left( \frac{\partial L}{\partial a}\frac{\partial a}{\partial o} + \frac{\partial L}{\partial b}\frac{\partial b}{\partial o} + \frac{\partial L}{\partial c}\frac{\partial c}{\partial o} + \frac{\partial L}{\partial d}\frac{\partial d}{\partial o} \right) \\ \nonumber
&~~+ \textit{k}~\left( \frac{\partial L}{\partial a}\frac{\partial a}{\partial p} + \frac{\partial L}{\partial b}\frac{\partial b}{\partial p} + \frac{\partial L}{\partial c}\frac{\partial c}{\partial p} + \frac{\partial L}{\partial d}\frac{\partial d}{\partial p} \right) \\ \nonumber
&= \frac{\partial L}{\partial a} \left( \frac{\partial a}{\partial m} + \textit{i}~\frac{\partial a}{\partial n} + \textit{j}~\frac{\partial a}{\partial o} + \textit{k}~\frac{\partial a}{\partial p} \right) \\ \nonumber
&~~+ \frac{\partial L}{\partial b} \left( \frac{\partial b}{\partial m} + \textit{i}~\frac{\partial b}{\partial n} + \textit{j}~\frac{\partial b}{\partial o} + \textit{k}~\frac{\partial b}{\partial p} \right) \\ \nonumber
&~~+ \frac{\partial L}{\partial c} \left( \frac{\partial c}{\partial m} + \textit{i}~\frac{\partial c}{\partial n} + \textit{j}~\frac{\partial c}{\partial o} + \textit{k}~\frac{\partial c}{\partial p} \right) \\ \nonumber
&~~+ \frac{\partial L}{\partial d} \left( \frac{\partial d}{\partial m} + \textit{i}~\frac{\partial d}{\partial n} + \textit{j}~\frac{\partial d}{\partial o} + \textit{k}~\frac{\partial d}{\partial p} \right) \\ \nonumber
&= \frac{\partial L}{\partial \mathbb{R}(q)} \left( \frac{\partial a}{\partial m} + \textit{i}~\frac{\partial a}{\partial n} + \textit{j}~\frac{\partial a}{\partial o} + \textit{k}~\frac{\partial a}{\partial p} \right) \\ \nonumber
&~~+ \frac{\partial L}{\partial \mathbb{I}(q)} \left( \frac{\partial b}{\partial m} + \textit{i}~\frac{\partial b}{\partial n} + \textit{j}~\frac{\partial b}{\partial o} + \textit{k}~\frac{\partial b}{\partial p} \right) \\ \nonumber
&~~+ \frac{\partial L}{\partial \mathbb{J}(q)} \left( \frac{\partial c}{\partial m} + \textit{i}~\frac{\partial c}{\partial n} + \textit{j}~\frac{\partial c}{\partial o} + \textit{k}~\frac{\partial c}{\partial p} \right) \\ \nonumber
&~~+ \frac{\partial L}{\partial \mathbb{K}(q)} \left( \frac{\partial d}{\partial m} + \textit{i}~\frac{\partial d}{\partial n} + \textit{j}~\frac{\partial d}{\partial o} + \textit{k}~\frac{\partial d}{\partial p} \right) \\ \nonumber
&= \mathbb{R}(\nabla_L(q)) \left( \frac{\partial a}{\partial m} + \textit{i}~\frac{\partial a}{\partial n} + \textit{j}~\frac{\partial a}{\partial o} + \textit{k}~\frac{\partial a}{\partial p} \right) \\ \nonumber
&~~+ \mathbb{I}(\nabla_L(q)) \left( \frac{\partial b}{\partial m} + \textit{i}~\frac{\partial b}{\partial n} + \textit{j}~\frac{\partial b}{\partial o} + \textit{k}~\frac{\partial b}{\partial p} \right) \\ \nonumber
&~~+ \mathbb{J}(\nabla_L(q)) \left( \frac{\partial c}{\partial m} + \textit{i}~\frac{\partial c}{\partial n} + \textit{j}~\frac{\partial c}{\partial o} + \textit{k}~\frac{\partial c}{\partial p} \right) \\ \nonumber
&~~+ \mathbb{K}(\nabla_L(q)) \left( \frac{\partial d}{\partial m} + \textit{i}~\frac{\partial d}{\partial n} + \textit{j}~\frac{\partial d}{\partial o} + \textit{k}~\frac{\partial d}{\partial p} \right)
\label{eq:diff2}
\end{align}

\subsection{Whitening a Matrix}\label{a:whitening}
Let $\textbf{X}$ be an $n$~x~$n$ matrix and $\mbox{cov}(\textbf{X}) = \mathbf{\Sigma}$ is the symmetric covariance matrix of the same size.
Whitening a matrix linearly decorrelates the input dimensions, meaning that whitening transforms $\textbf{X}$ into $\textbf{Z}$ such that $\mbox{cov}(\textbf{Z}) = \textbf{I}$ where $\textbf{I}$ is the identity matrix \cite{kessy2017optimal}. 
The matrix $\textbf{Z}$ can be written as:
\begin{equation}
\textbf{Z} = \textbf{W}(\textbf{X} - \mu)
\label{eq:white1}
\end{equation}
where $\textbf{W}$ is an $n$~x~$n$ `whitening' matrix. Since $\mbox{cov}(\textbf{Z}) = \textbf{I}$ it follows that:
\begin{align}
&\mathbb{E}[\mathbf{Z}\mathbf{Z}^T] = \mathbf{I} \nonumber \\
&\mathbb{E}[\mathbf{W}(\mathbf{X - \mu})(\mathbf{W}(\mathbf{X} - \mu))^T] = \mathbf{I} \nonumber \\
&\mathbb{E}[\mathbf{W}(\mathbf{X - \mu})(\mathbf{X} - \mu)^T\mathbf{W}^T] = \mathbf{I} \nonumber \\
&\mathbf{W}\Sigma\mathbf{W}^T = \mathbf{I} \nonumber \\
&\mathbf{W}\Sigma\mathbf{W}^T\mathbf{W} = \mathbf{W} \nonumber \\
&\mathbf{W}^T \mathbf{W} = \mathbf{\Sigma}^{-1} \label{eq:white2}
\end{align}
From \eqref{eq:white2} it is clear that the Cholesky decomposition provides a suitable (but not unique) method of finding $\textbf{W}$.

\subsection{Cholesky Decomposition}
Cholesky decomposition is an efficient way to implement LU decomposition for symmetric matrices, which allows us to find the square root.
Consider $\textbf{A}\textbf{X} = \textbf{b}$, $\textbf{A}=[a_{ij}]_{n\times n}$, and $a_{ij} = a_{ji}$, then the Cholesky decomposition of $\textbf{A}$ is given by $\textbf{A} = \textbf{L}\textbf{L}'$ where
\begin{equation}
\textbf{L}=
\begin{bmatrix}
 l_{11} & 0 & \ldots & 0 \\
 l_{21} & l_{22} & \ldots & \vdots \\
 \vdots & \vdots & \ddots & 0 \\
 l_{n1} & l_{n2} & ... & l_{nn} \\
\end{bmatrix}
\label{eq:cholesky1}
\end{equation}
Let $l_{ki}$ be the $k^{th}$ row and $i^{th}$ column entry of $\textbf{L}$, then

\[ 
   l_{ki} = 
	 \begin{cases} 
      0, & k < i \\
      \sqrt{a_{ii} - \sum_{j=1}^{i-1}l^2_{kj}}, & k=i \\
      \frac{1}{l_{ii}} (a_{ki} - \sum_{j=1}^{i-1}l_{ij}l_{kj}), & i < k 
   \end{cases}
\]

\subsection{4 DOF Independent Normal Distribution}
Consider the four-dimensional vector $\textbf{Y} = (S,T,U,V)$ which has components that are normally distributed, centered at zero, and independent. 
Then $S$, $T$, $U$, and $V$ all have density functions
\begin{equation}
f_S(x;\sigma) = f_T(x;\sigma) = f_U(x;\sigma) = f_V(x;\sigma) = \frac{e^{-x^2/(2\sigma^2)}}{\sqrt{2\pi\sigma^2}}.
\label{eq:single_dists}
\end{equation}
Let $\textbf{X}$ be the length of $\textbf{Y}$, which means $\textbf{X} = \sqrt{S^2+T^2+U^2+V^2}$.
Then $\textbf{X}$ has the cumulative distribution function
\begin{equation}
F_X(x;\sigma) = \int \!\!\!\int \!\!\!\int \!\!\!\int_{H_x} \!\!f_S(\mu;\sigma)f_T(\mu;\sigma)f_U(\mu;\sigma)f_V(\mu;\sigma) ~dA,
\label{eq:cumdist}
\end{equation}
where $H_x$ is the four-dimensional sphere
\begin{equation}
H_x = \left\{(s,t,u,v)~:~\sqrt{s^2+t^2+u^2+v^2} < x \right\}.
\label{eq:4dsphere}
\end{equation}
We then can write the integral in polar representation
\begin{align}
F_X(x;\sigma) = & ~\frac{1}{4\pi^2\sigma^4} \!\int_0^\pi \!\!\!\!\int_0^\pi \!\!\!\!\int_0^{2\pi} \!\!\!\!\!\int_0^x \!r^3e^{\frac{-r^2}{2\sigma^2}} \mbox{sin}(\theta) \mbox{sin}(\phi) \mbox{cos}(\psi) ~dr d\theta d\phi d\psi \nonumber \\
= & ~\frac{1}{2\sigma^4} \int_0^x r^3e^{-r^2/(2\sigma^2)} ~dr.
\label{eq:polarint}
\end{align}
The probability density function of $\textbf{X}$ is the derivative of its cumulative distribution function so we use the funamental theorem of calculus on \eqref{eq:polarint} to finally arrive at
\begin{equation}
f_X(x;\sigma) = \frac{d}{dx}F_X(x;\sigma) =  ~\frac{1}{2\sigma^4} x^3e^{-x^2/(2\sigma^2)}.
\label{eq:finaldist}
\end{equation}

\end{document}